\documentclass[10pt,twocolumn,letterpaper]{article}

\usepackage{cvpr}
\usepackage{times}
\usepackage{epsfig}
\usepackage{graphicx}
\usepackage{amsmath}
\usepackage{amssymb}

\usepackage[breaklinks=true,bookmarks=false]{hyperref}

\cvprfinalcopy % *** Uncomment this line for the final submission

\def\cvprPaperID{****} % *** Enter the CVPR Paper ID here

\begin{document}

%%%%%%%%% TITLE
\title{Joint Deep Learning for Car Detection}

\author{Seyedshams Feyzabadi\\
Electrical Engineering and Computer Science\\
University of California, Merced\\
{\tt\small sfeyzabadi@ucmerced.edu}
% For a paper whose authors are all at the same institution,
% omit the following lines up until the closing ``}''.
% Additional authors and addresses can be added with ``\and'',
% just like the second author.
% To save space, use either the email address or home page, not both
%{\tt\small secondauthor@i2.org}
}

\maketitle
%\thispagestyle{empty}

%%%%%%%%% ABSTRACT
\begin{abstract}
Traditional object recognition approaches apply feature extraction, part deformation handling, occlusion handling and classification  sequentially while they are independent from each other. Ouyang and Wang proposed a model for jointly learning of all of the mentioned processes using one deep neural network. We utilized, and manipulated their toolbox in order to apply it in car detection scenarios where it had not been tested. Creating a single deep architecture from these components, improves the interaction between them and can  enhance the performance of the whole system. We believe that the approach can be used as a general purpose object detection toolbox. We tested the algorithm on UIUC car dataset, and achieved an outstanding result. The accuracy of our method was 97 \% while the previously reported results showed an accuracy of up to 91 \%. We strongly believe that having an experiment on a larger dataset can show the advantage of using deep models over shallow ones. 
\end{abstract}

%%%%%%%%% BODY TEXT
\section{Introduction}

Object recognition plays an important role in computer vision, and has been extensively discussed in the literature \cite{lowe1999object, belongie2002shape, riesenhuber1999hierarchical, mohan2001example}.  The main difficulty in creating a robust object detection approach comes from the wide range of variations in images of objects belonging to the same object class. Similar objects from the same class might have different sizes, different shapes because of the view points etc. Having a part-based model helps to overcome these issues and create an abstract model of the object by using its parts. Similar challenges of different textures, lighting, background and etc. increase the difficulties of detecting an object in a picture.

The traditional approach of overcoming these challenges follows the following steps:
\begin{enumerate}
\item Feature extraction: features are extracted from an image to find out the most discriminative parts of it. The most famous approaches are SIFT \cite{ng2003sift} or Haar \cite{viola2003detecting}
\item Deformation handling: extracting multiple parts of an object will be very helpful when there are variety of shapes for the same object while they consist of similar part. For instance, cars have different shapes, but they all consist of body, tires, trunk and hood. Agarwal performed a good research on part-based object detection  \cite{agarwal2004learning, agarwal2002learning}.
\item Occlusion handling: Occluded parts have to be detected in order to avoid using them in object detection itself. Including them in the detection phase will reduce the accuracy of the program. Some usages of occlusion detection algorithms can be found here \cite{wang2009hog, enzweiler2010multi}.
\item Classification: After finishing all the procedures, a classifier decides whether the chosen window contains an instance of that object or no. The most known approaches are  SVM \cite{dalal2005histograms} and random forests \cite{dollar2012crosstalk}.
\end{enumerate}

Most of the traditional approaches follow these steps sequentially. Each of them is created and trained first, and the output of each step is fed as the input to the next step. However, the interactions and feedbacks of each step to the other one has not been studied well.   

\begin{figure*}[htb]
\begin{center}
   \includegraphics[width=\linewidth]{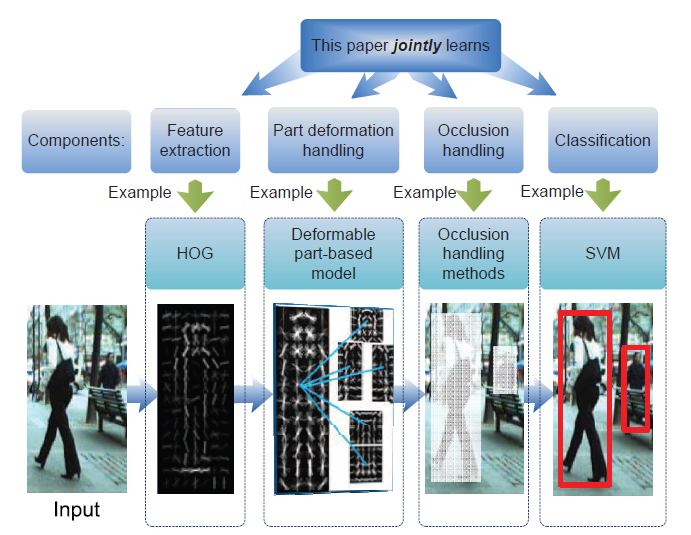}
\end{center}
   \caption{Joint deep learning for pedestrian detection, taken from  \cite{ouyang2013joint}}
\label{fig:joint}
\end{figure*}

Ouyang et al \cite{ouyang2013joint} proposed the first model for joint learning of all of these steps in one single deep convolutional  network and reported good results in pedestrian detection scenarios. Their model is based on learning all of the steps at the same time, and use the output of higher levels in training the lower levels. Figure \ref{fig:joint}  shows their approach for joint deep learning in pedestrian detection. Their method is capable of using the training data in creating better low level filters. This vividly explains why the joint learning can be beneficial comparing to independent learning. 

In this paper we focus on Ouyang's work \cite{ouyang2013joint} for pedestrian detection and mostly base our work on their efforts. The main contribution of this paper is to generalize their framework and apply it on other applications. The author strongly believes that this approach has the potential to become a generic framework for object recognition. The biggest advantage of this approach is its flexibilities where it makes the approach very easy to change application. We can imagine that the same method can be applied to other objects too. We also believe that the deformation layer of this methodology makes it more useful for objects that are composed of many parts. As an example a bicycle can be a good representative object. Another interesting application can be in animal detection where their body parts are not rigid and may have different articulations.

%\begin{table}
%\begin{center}
%\begin{tabular}{|l|c|}
%\hline
%Method & Frobnability \\
%\hline\hline
%Theirs & Frumpy \\
%Yours & Frobbly \\
%Ours & Makes one's heart Frob\\
%\hline
%\end{tabular}
%\end{center}
%\caption{Results.   Ours is better.}
%\end{table}
\section{Related Work}
Several approaches to object detection were proposed in the past that use some form of learning. In most such approaches, images are represented by using some features, and a learning method is used to identify regions in the feature space that correspond to the object class. There are a large
variety in types of features used and the learning methods applied \cite{forsyth2002computer}.

Recently deep models showed great performance in machine learning and recognition problems \cite{krizhevsky2012imagenet, hinton2006fast, jarrett2009best, le2013building, ranzato2007unsupervised, ess2007depth} comparing to shallow approaches. There are some robotics usages of deep and shallow models too\cite{shams1, shams2, shams3}. They proved and showed that these approaches are capable of solving complex tasks. Their complicated structure makes them more compatible for complex problems with a lot of details. But, it arises the problem of unknown regions. There is not any proof why they are so efficient \cite{krizhevsky2012imagenet}, how they can be improved or what is a structured way of creating the best model. 

Most of the traditional approaches use features such as SIFT or HOG \cite{dalal2005histograms} to extract the overall shape of an object, and apply different learning approaches to train the system. All of these approaches have one common point that the features are manually generated, and only used for the training purposes. While some recent investigations show that learning features from the training data is very helpful and improves the accuracy of the program \cite{bar2010part}.  

Detecting cars is an important aspect in traffic control scenarios. It will be very useful to distinguish cars from other objects. Multiple approaches have focused on this problem \cite{goerick1996artificial, agarwal2004learning, agarwal2002learning}. Since different types of cars have different shapes, but are composed from similar parts, it will be very beneficial to create a part-based model for it. Then the same solver can be applied to different classes of cars such as sedans, SUVs, trucks and etc. Comparing to other objects, cars have very huge variety of shapes and sizes. Thus it will be too hard, if not impossible, to have a general model for all of them. But in this work, we only focus on sedan cars.  

We propose a model for car detection that benefits from deep learning approaches and is capable of detecting different classes of cars. The approach uses the training data to improve its low-level features. Even though we applied this approach to side view pictures of cars, it is extendable to other views of the cars too. We very closely followed the work by Ouyang \cite{ouyang2013joint} and used the same methodology to detect different object class.

\section{Our Approach}

\begin{figure*}[htb]
\begin{center}
   \includegraphics[width=\linewidth]{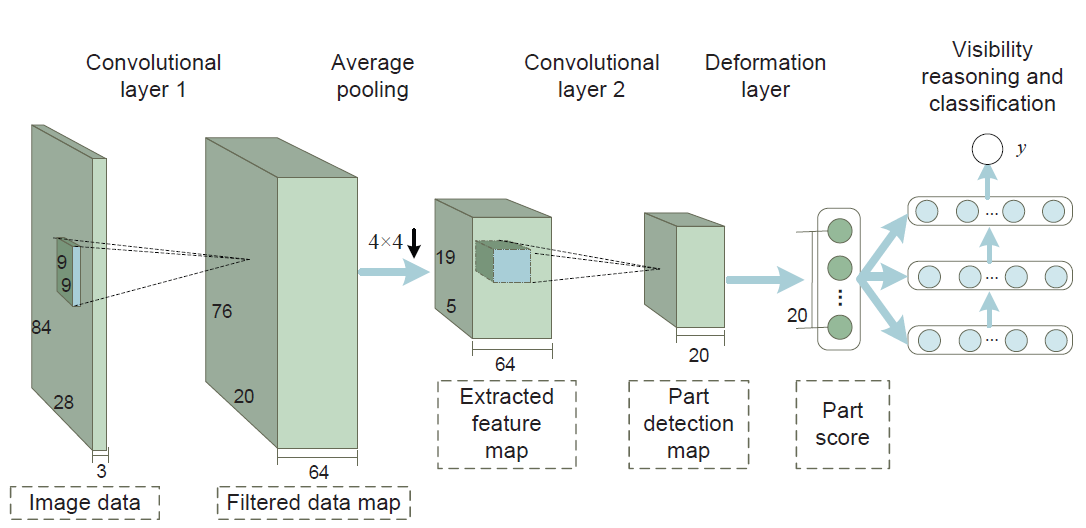}
\end{center}
   \caption{Program architecture, taken from \cite{ouyang2013joint}}
\label{fig:architecture}
\end{figure*}

As discussed earlier, the main difference between this method with traditional recognition method is the jointly learning of the mentioned processes. Its architecture is shown in Figure \ref{fig:architecture}. In general, the following steps are performed in this program:
\begin{itemize}
\item The first convolutional layer generates filtered data maps. This layer convolves the 3-channel input image data with 9 x 9 x 3 filters and outputs 64 maps.
\item Average pooling is applied to Features maps with the size of 64 filtered data maps using 4 x 4 boxcar filters with a 4 x 4 subsampling step.
\item The second convolutional layer generates part detection maps. This layer convolves the feature maps with 8 part filters of different sizes and outputs 8 part detection maps.
\item Part scores are obtained from the 8 part detection maps using a deformation handling layer. This layer outputs 8 part scores.
\item The visibility reasoning of 8 parts is used for estimating the label y; that is, whether a given window encloses a car or not.
\end{itemize}

\subsection{Input Data}
The size of an input image is assumed to be 84x28 and it is fed to the program with the following three channels:
\begin{itemize}
\item The first contains Y data in YUV format. Its size is  84 x 28.
\item The three-channel 42 x 14 images in the YUV color space are appended into the second channel of size 84 x 28 with zero padding.
\item Four 42 x 14 edge maps are appended into the third channel of size 84x28. Three edge maps are obtained from the three-channel images in the YUV color space. The magnitudes of horizontal and vertical edges are computed using the Sobel edge detector \cite{jahne1999principles}. The fourth edge map is chosen as the  maximum magnitudes from the first three edge maps.
\end{itemize}

Ouyang et al, found it useful to feed the program in three channels as explained above, and we follow their strategy. They claimed that \begin{quote} 
In this way, information about pixel values at different
resolutions and information of primitive edges are utilized
as the input of the first convlutional layer to extract features.
The first convolutional layer and its following average pooling
layer use the standard CNN settings.
We empirically find that it is better to arrange the images and edge maps into three concatenated channels instead of eight separate channels. In order to deal with illumination change, the data in each channel is preprocessed to be zero mean and unit variance \cite{ouyang2013joint}.\end{quote}

\subsection{Part Maps}
\begin{figure}[htb]
\begin{center}
   \includegraphics[width=\linewidth]{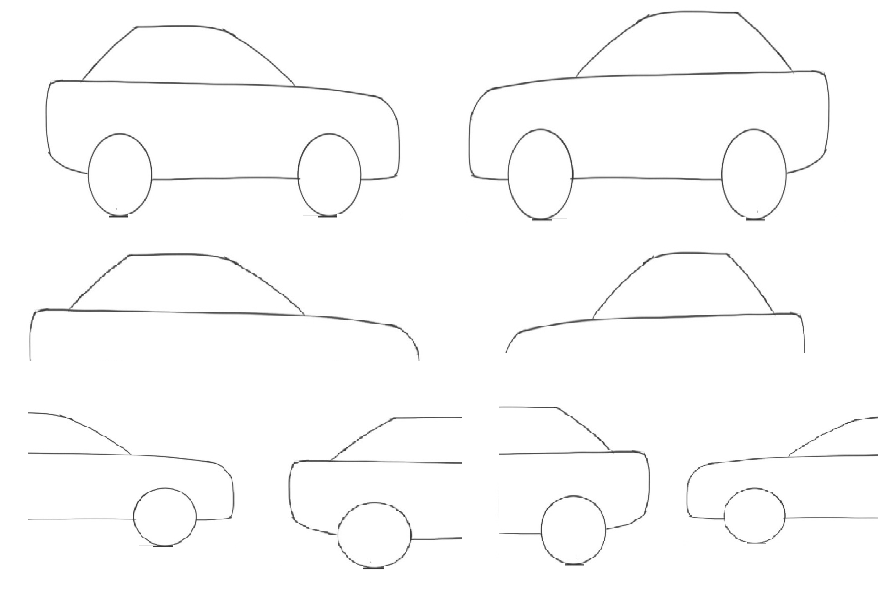}
\end{center}
   \caption{Part filters for the side view pictures of cars. }
\label{fig:filters}
\end{figure}

Most of the vision approaches have filters with equal sizes, but cars might have different sizes or been seen from different distances.. Therefore, we decided to have variable size filter as shown in Figure \ref{fig:filters}. The filters are defined on three levels and the size of the largest one is 5 x 15. The smaller ones have portion of the whole filter. The side view image of a car can be taken from both sides, therefore, it is needed to have filters for both views. The left and right side of Figure \ref{fig:filters} deal with this issue.

\subsection{Deformation Layer}
In this subsection we would like to explain what happens in the second convolutional layer. 
This layer receives the P part detection maps as input and  part scores s = \{$s_1, . . . , s_P $\}, P = 8, are its output. The convolutional layer treats the detection maps individually and produces the pth part score $s_p$ from the pth part detection map, denoted by $M_p$. A 2D summation map,  $B_p$, is calculated by summing up the part detection map $M_p$ and the deformation maps as follows:
\[ B_p = M_p + \Sigma c_{n,p}D_{n,p} \]

$D{n,p}$ denotes the nth deformation map for the pth part, $c_{n,p}$
denotes the weight for $D_{n,p}$, and N denotes the number of
deformation maps. sp is globally max-pooled from $B_p$
\[s_p = max b_p^{(x,y)} \]

where $b_p^{(x,y)}$
 denotes the (x, y)th element of $B_p$. The detected
part location can be inferred from the summed map
as follows:

\[(x,y)_p = arg max b_p^{(x,y)}\]

At the training stage, only the value at location $(x, y)_p$ of
$B_p$ is used for learning the deformation parameters.

The $c_{n,p}$ and $D_{n,p}$ are the key for designing different
deformation models. Both $c_{n,p}$ and $D_{n,p}$ can be considered
as the parameters to be learned. Three examples are
given below:
\begin{enumerate}
\item Suppose N = 1, $c_{1,p}$ = 1 and the deformation
map $D_{1,p}$ is to be learned. In this case, the discrete
locations of the pth part are treated as bins and the deformation
cost for each bin is learned. $d_{1,p }^{(x,y)}$, which denotes
the (x, y)th element of D1,p, corresponds to the deformation
cost of the pth part at location (x, y).
\item $D_{1,p}$ can also be predefined. Suppose N = 1
and $c_{n,p}$ = 1. If $d_{1,p}^{(x,y)}$
 is the same for any (x, y), then there
is no deformation cost. If $d_{1,p}^{(x,y)} = - \infty$
for $(x, y) \in X$, $d_{1,p}^{(x,y)} = 0$
for $(x, y) \in X$, then the parts are only allowed to
move freely in the location set X. Max-pooling is a special case of this example by setting X to be a local region. The
disadvantage of max-pooling is that the hand-tuned local
region does not adapt to different deformation properties of
different parts.
\item The deformation layer can represent the
widely used quadratic constraint of deformation. Below,
we skip the subscript p.
The quadratic constraint of deformation can be represented
as follows:
\begin{equation}
b^{(x,y)} = m^{(x,y)} + c_1(x - a_x + \frac{c_3}{2c_1})^2 + c_2(y - a_y + \frac{c_4}{2c_2})^2
\end{equation}
where $m^{(x,y)}$ is the (x, y)th element of the part detection
map M, $(a_x, a_y)$ is the predefined anchor location of the
pth part. They are adjusted by c3/2c1 and c4/2c2, which
are automatically learned. c1 and c2 decide the deformation
cost. There is no deformation cost if c1 = c2 = 0.
Parts are not allowed to move if $c1 = c2 = −\infty$. (ax, ay)
and $(\frac{c_3}{2c_1}, \frac{c_4}{2c_2})$ jointly decide the center of the part. The
quadratic constraint can be represented as follows:
\begin{align}
B =  M + & c_1D_1 + c_2D_2 + c_3D3 + c_4D_4 + c_5.1 , \\
b^{(x,y)} = & m^{(x,y)} + c_1d_1^{(x,y)} + c_2d_2^{(x,y)}  + c_3d_3^{(x,y)} \nonumber \\
		&  + c_4d_4^{(x,y)} + c_5,  \nonumber \\
d_1^{(x,y)} = & (x - a_x)^2, \nonumber \\
d_2^{(x,y)} =  & (y - a_y)^2, \nonumber \\
d_3^{(x,y)} = & x - a_x, \nonumber \\
d_4^{(x,y)} = & y - a_y, \nonumber \\
c_5 = & \frac{c_3^2}{4c_1} + \frac{c_4^2}{4c_2}\nonumber 
\end{align}
where $1$ is a matrix with all elements being one, $d_n^{(x,y)}$
 is the (x, y)th element of $D_n$. In this case, c1, c2, c3 and c4 are parameters
to be learned and $D_n$ are predefined. c5 is the same in all
locations and need not be learned.
\end{enumerate}

\begin{figure}[htb]
\begin{center}
   \includegraphics[width=\linewidth]{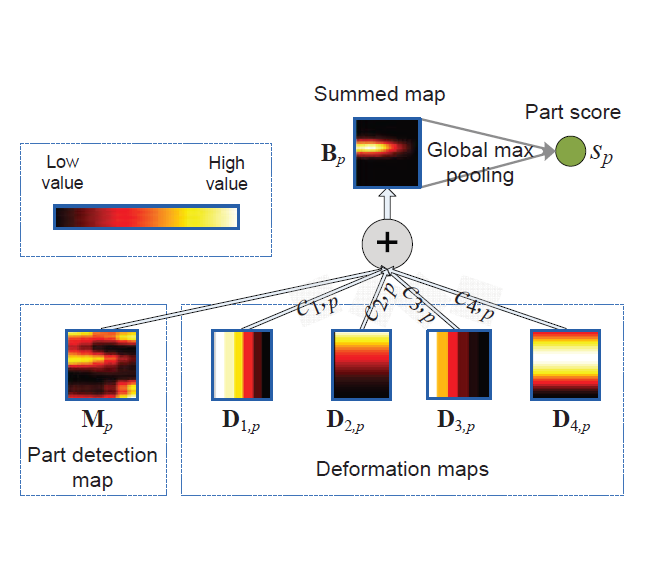}
\end{center}
   \caption{The deformation layer. Part detection map and deformation maps are summed up
with weights $c_{n,p}$ for n = 1, 2, 3, 4 to obtain the summed map
$B_p$. Global max pooling is then performed on the summed map to
obtain the score sp for the pth part. Taken from \cite{ouyang2012discriminative}.}
\label{fig:scores}
\end{figure}

Figure \ref{fig:scores} is an example of learning these parameters. In this case parameters $c_1$, $c_2$, $c_3$ and  $c_4$ are to be learned and $D_n$ is predefined. It is notable that a large part of this subsection follows the work of Ouyang.

\subsection{Classification}

The results of previous section is a set of scores $s_1, s_2, ..., s_n$. Then visibility reasoning is used for classification. More details of it can be found in \cite{ouyang2012discriminative}. 

In this paper, features, deformable
models, and visibility relationships are jointly learned.
Backpropagation through $s$ is used as prediction in order to learn the parameters in the two convolutional layers
and the deformation layer in Figure \ref{fig:architecture}.

In order to train this deep architecture, we used a multi-stage
training strategy. We start with a 1-layer CNN using supervised
training. Since Gabor filters \cite{jain1990unsupervised} are similar to the human visual system,
they are used for initializing the first CNN. We add one more
layer at each stage, the layers trained in the previous stage are
used for initialization and then all the layers at the current stage
are jointly optimized with back propagation.

\section{Experimental Results}

\begin{figure*}[htb]
\begin{center}
   \includegraphics[width=\linewidth]{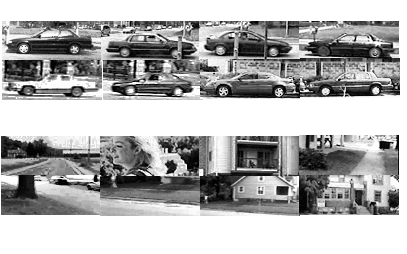}
\end{center}
   \caption{UIUC dataset. The upper part shows the positive images, and the lower part is the negative samples.}
\label{fig:dataset}
\end{figure*}

The described architecture is tested on UIUC database \footnote{\url{http://cogcomp.cs.illinois.edu/Data/Car/}}. The database contains 1050 images(550 cars and 500 non-cars). The test set consists of 170 images with 200 cars in them. The file format of the images are PGM. There are some evaluation codes in the dataset which we did not use for this experiment, but they can be utilized for evaluating different object recognition algorithms. A snapshot of the data set is shown in Figure \ref{fig:dataset}.

Image sizes are 40 x 100 and they are in gray scale. In order to be in line with Ouyang's work, we rotated and scaled the images to 84X28. Therefore, we can use the same set of filter as them. 

We used log average miss rate in order to make comparisons with other methods. It is computed by averaging the miss rate
at nine FPPI rates that are evenly spaced in the log-space in the range from 0.01 to 1. 

In order to improve the results we applied augmentation. This is done by creating manual rotations on the images starting from -10 degrees up to +10 with intervals of 1 degree.

\begin{figure}[htb]
\begin{center}
   \includegraphics[width=\linewidth]{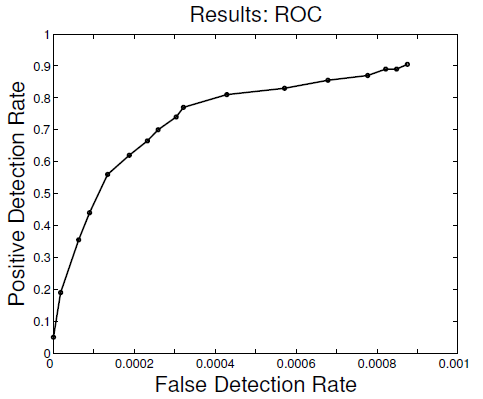}
\end{center}
   \caption{The results of Agarwal's work, published in \cite{agarwal2004learning}. }
\label{fig:uiuc}
\end{figure}

As a comparison method we picked the method developed by Agarwal et al \cite{agarwal2004learning}. They used the same data set with similar test images. Their method is robust and accurate and a sample results of their method is shown in Figure \ref{fig:uiuc}. Their average miss rate is 9 \%.

\begin{figure}[htb]
\begin{center}
   \includegraphics[width=\linewidth]{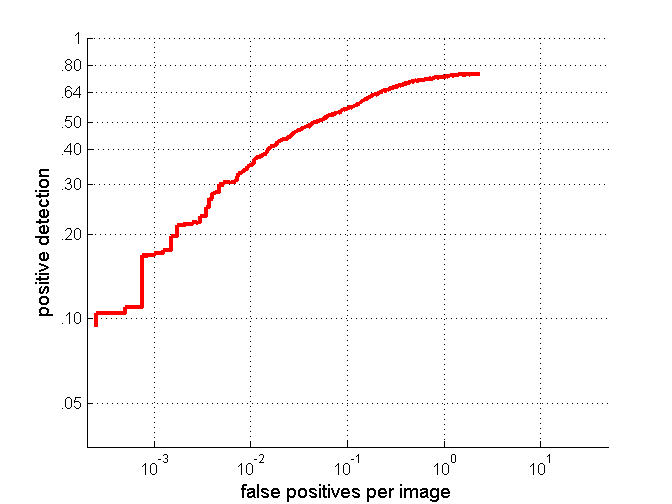}
\end{center}
   \caption{Jointly learning without Augmentation.}
\label{fig:noaug}
\end{figure}

The results of our experiments without augmentation is shown in Figure \ref{fig:noaug}. However, the results is not disappointing. It is not too bad either. The miss rate of this approach is 23 \%.

\begin{figure}[htb]
\begin{center}
   \includegraphics[width=\linewidth]{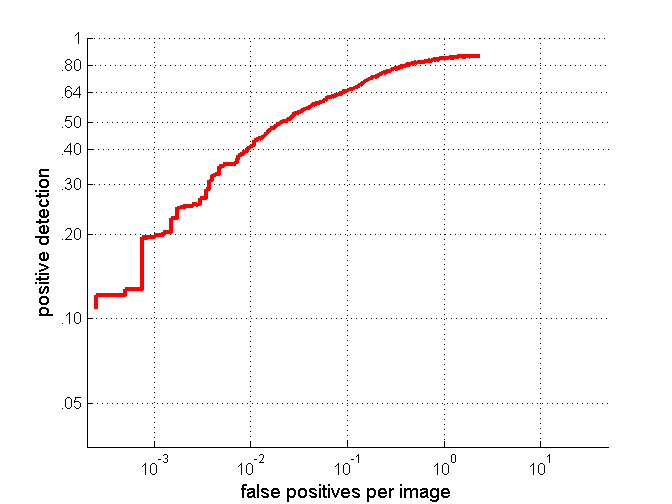}
\end{center}
   \caption{The results of our algorithm after applying augmentation.}
\label{fig:aug}
\end{figure}

Augmentation is needed to improve the dataset and makes it more stable against camera angles. After we applied the augmentation, we achieved the results in Figure \ref{fig:aug}. This method decreased the average miss rate to 3 \% which shows a great improvement comparing to the reported 9 \% from Agarwal's work.

\section{Conclusions}
In this report we showed our work about using joint deep learning architecture for car detection applications. The main idea of joint deep learning is to include feature extraction, part deformation handling, occlusion handling and classification in one single deep neural network where two layers of convolutional layers exist. The first layer extracts feature maps using low-level features that are tuned during the training phase. And the second convolutional layer is responsible for deformation handling to extract a score based on different visible parts of an object. 

We applied the algorithm on UIUC car dataset, and managed to gain good results of up to 97 \% of success rate, after applying augmentation. Comparing our results with older results of Agarwal, our method outperforms the Agarwal's result by  6\%. We also showed the importance of augmentation in the final performance of an algorithm. Since deep learning approaches require large datasets, we increased the size of our database by applying augmentation. 

{\small
\bibliographystyle{ieee}
\bibliography{egbib}
}

\end{document}